\documentclass[submission,copyright,creativecommons]{eptcs}
\usepackage[utf8]{inputenc}
\usepackage[T1]{fontenc}
\usepackage{textcomp}
\usepackage[english]{babel}
\usepackage{amsmath, amssymb}
\usepackage{tikz}
\usepackage{xfp, siunitx}
\usepackage{pgfplots}

\usepackage{gb4e}
\noautomath 

\usepackage{import}
\usepackage{xifthen}
\usepackage{pdfpages}
\usepackage{transparent}

\newtheorem{definition}{Definition}

\usetikzlibrary{arrows,positioning,matrix,backgrounds,calc,fit,decorations.pathreplacing,intersections,through}
\usepgfplotslibrary{statistics, fillbetween}
\pgfplotsset{compat=newest}
\pgfplotsset{height=4cm, width=7cm}
\pgfplotsset{
  layers/axis lines on top/.define layer set={
    axis background,
    axis grid,
    axis ticks,
    axis tick labels,
    pre main,
    main,
    axis lines,
    axis descriptions,
    axis foreground,
  }{/pgfplots/layers/standard},
}

\newcommand{\histogram}[3][]{
\begin{tikzpicture}
\begin{axis}[
    ymin=0,
    xmin=0, xmax=1,
    xtick={0,1/6,2/6,3/6,4/6,5/6,1},
    xticklabels={$0$,  $\frac{1}{6}$,  $\frac{2}{6}$,  $\frac{3}{6}$,  $\frac{4}{6}$,  $\frac{5}{6}$,  $1$},
    xlabel={Signalling fraction},
    minor y tick num = 4,
    grid,
    grid style=dashed,
    legend pos={outer north east},
    legend cell align={left},
    ]
\addplot+[ybar interval,mark=no,color=blue,fill=blue,fill opacity=0.3,ybar legend] plot coordinates {
#2 (0.16666667, 0)
};
\addplot+[ybar interval,mark=no,color=red,fill=red,fill opacity=0.3,ybar legend] plot coordinates {
#3 (1, 0)
};
\ifthenelse{\isempty{#1}}
{\legend{contextual, non-contextual}}
{}
\end{axis}
\end{tikzpicture}
}

\newcommand{\textexample}[1]{
  \begin{exe}
    \ex #1      
  \end{exe}
}

\newcommand{\cnt}[1]{{\textsf{CNT}\text{$_#1$}}}
\newcommand{\cf}{\textsf{CF}}

\begin{document}
\title{Generalised Winograd Schema and its Contextuality}
\def\titlerunning{Winograd Schema, Contextuality, Sheaves}
\providecommand{\event}{QPL 2023} 

\author{Kin Ian Lo \qquad Mehrnoosh Sadrzadeh
\institute{University College London \\London, UK}
\email{\{kin.lo.20,m.sadrzadeh\}@ucl.ac.uk}
\and
Shane Mansfield
\institute{Quandela \\ Paris, France}
\email{shane.mansfield@quandela.com}
}

\def\authorrunning{K. I. Lo, et al.}

\maketitle 

Ambiguities in natural language give rise to probability distributions over interpretations. The distributions are often over multiple ambiguous words at a time; a multiplicity which makes them a suitable topic for sheaf-theoretic models of quantum contextuality. 
Previous research showed that different quantitative measures of contextuality correlate well with Psycholinguistic research on lexical ambiguities.
In this work, we focus on coreference ambiguities and investigate the Winograd Schema Challenge (WSC), a test proposed by Levesque in 2011 to evaluate the intelligence of machines.
The WSC consists of a collection of multiple-choice questions that require disambiguating pronouns in sentences structured according to the Winograd schema, in a way that makes it difficult for machines to determine the correct referents but remains intuitive for human comprehension.
In this study, we propose an approach that analogously models the Winograd schema as an experiment in quantum physics.
However, we argue that the original Winograd Schema is inherently too simplistic to facilitate contextuality. 
We introduce a novel mechanism for generalising the schema, rendering it analogous to a Bell-CHSH measurement scenario.
We report an instance of this generalised schema, complemented by the human judgements we gathered via a crowdsourcing platform.
The resulting model violates the Bell-CHSH inequality by 0.192, thus exhibiting contextuality in a coreference resolution setting.
\section{Introduction}

The Winograd Schema Challenge (WSC) originated from the ideas of the American computer scientist Terry Winograd in the 1970s. Winograd was interested in situations where machine understanding could fall behind human understanding. 
He constructed hypothetical experiments where humans and machines would read a given description, and then answer some questions about it. 
The descriptions would provide humans with enough context and thus they could answer the questions correctly. 
However, machine understanding would fall short, as machines did not learn from the context in the same way as humans did. 
An example description is the sentence ``The city councilmen refused the demonstrators a permit because they feared violence.''. The question following it is ``Who feared violence?'' and the correct answer is ``The city councilmen''. If we change the word ``feared'' to ``advocated'', the question will have the opposite answer, namely ``the demonstrators''. 
Winograd's examples were picked up by the Canadian AI scientist Hector Levesque in 2011. 
He created a suite of descriptions and questions, proposing them as a test of machine intelligence - an alternative to the Turing Test \cite{Levesque2012}.
Later, the AI company Nuance put forwards a cash prize of USD~25,000 for any AI that could solve the challenge with an accuracy close to humans, 92-96\%. 
No AI system managed to achieve the target, and as a result, the prize was withdrawn in 2018. It was not until the 2020s that large pre-trained language models, employing transformer architectures, eventually reached a performance level comparable to human accuracy \cite{Kocijan2022}. Despite these advancements, the WSC continues to present significant challenges for AI systems lacking extensive data resources and computational power.

In previous work, we showed how natural language notions of context can be modelled by the mathematics of quantum contextuality \cite{Wang2021a,Wang2021b,Wang2018}. 
In particular, we modelled anaphoric context in \cite{Lo2022}. 
Inspired by the reliance of the WSC on anaphoric context, we decided to explore whether quantum contextuality could potentially provide a solution to the challenge.

Our initial examination found that the WSC in its original form lacked the complexity required to be of interest from a quantum contextuality standpoint. Upon modelling the WSC within the sheaf theoretic framework, it became evident that the scenario was too simplistic to exhibit contextuality, as the models derived from it were deterministic.

This motivated us to extend the schema and allow it to be non-deterministic such that it can, in principle, host contextuality. 
This was achieved by introducing additional linguistic context, namely, (1) two special words rather than one and (2) two ambiguous pronouns instead of one. Consequently, we obtained more observables and more measurement contexts, leading to a scenario that resembles the Bell-CHSH scenario.

The above outlines the first contribution of this paper. 
Our second contribution lies in the instantiation of our generalized Winograd Schema and the collection of human judgments via a crowdsourcing platform.
This allowed us to calculate the violation of the Bell-CHSH inequality and thereby establish the contextuality of our model, which was constructed based on human judgments. 
We also modelled the data using the Contextuality-by-Default (CbD) framework of contextuality and calculated a corresponding CbD degree of contextuality. 
It was found that our probabilistic model exhibited contextuality in both the Bell-CHSH and CbD models.

\section{Contextuality}
The origins of contextuality research can be traced back to 1935, with the work of Einstein, Podolsky, and Rosen (EPR)~\cite{EPR}. 
In their work, they posited that the quantum mechanical description of physics was incomplete when two spatially separated parties were permitted to make measurements on an entangled system.
A way of formalising such theories is in terms of hidden variables, which, if known, might fully determine the outcome that would result from any given measurement.
Bell's theorem~\cite{Bell1966, Bell1964} in the 1960s showed that no hidden-variable theory exists for quantum mechanics unless the measurement outcomes were allowed to be dependent on which other measurements are performed simultaneously. 
Around the same time, Kochen and Specker~\cite{Kochen1967} independently demonstrated that there exists a set of measurements in a 3-dimensional Hilbert space such that a non-contextual hidden-variable theory cannot exist, regardless of the state of the system.
These two results, collectively known as the Bell-Kochen-Specker theorem, showed that a hidden-variable theory for quantum mechanics must be contextual, providing some clarity to the debate on a more fundamental theory conforming to certain classical intuitions for quantum mechanics.
The first attempt at experimentally verifying Bell's inequality was performed by Aspect et al.~\cite{Aspect1982}, with the most recent ones closing all known loopholes in the earlier experiments~\cite{Giustina2015,Hensen2015,Shalm2015}.
Thus it has been established that quantum physics is vastly different from classical physics -- a description of quantum physics that agrees with our classical intuition must be contextual.

Other than the philosophical implications, contextuality has been shown to possess computational power through non-classical correlations.
Anders and Browne first showed that certain measurements on GHZ states can be used to lift a linear classical computation into a universal classical computation~\cite{Anders2009}; Raussendorf later showed that the probability of success of such computation is bounded by the degree of contextuality~\cite{Raussendorf2013}, as measured by the contextual fraction~\cite{Abramsky2011,Abramsky2017}.
Subsequent work by Howard et al.\ revealed that contextuality is an essential ingredient for \emph{magic state distillation}, a process that yields specific quantum states known as \emph{magic states}~\cite{Howard2014}.
The current most promising fault-tolerant quantum computing scheme, the surface code~\cite{Kitaev2003}, only permits fault-tolerant computation with a subset of quantum operations which can be efficiently simulated by classical computers. 
Via state injection, these magic states can be used with surface code to allow for fully fault-tolerant universal quantum computation.
Thus, one might argue that contextuality carries an intrinsic computational power that is absent in non-contextual systems.

A variety of frameworks for modelling contextuality have been developed. These including the sheaf-theoretic framework~\cite{Abramsky2011,Abramsky2012,Abramsky2017}, the Contextuality-by-Default (CbD) framework~\cite{Dzhafarov2013,Dzhafarov2015,Dzhafarov2016a}, the graph-theoretic framework~\cite{Cabello2014a}, a framework based on simplicial sets~\cite{Okay2022}. Generally speaking, these frameworks enable the formalisation of the notion of measurement through the use of various mathematical structures. Bell's inequalities, or in general inequalities that witness contextuality, can be derived systematically within these frameworks.
Although we will mainly use the terminology from the sheaf-theoretic framework to describe our examples, our results are framework-agonistic.

\subsection{Sheaf Theoretic Framework}
Here, we provide a concise overview of the sheaf-theoretic framework of contextuality proposed by Abramsky and Brandenburger~\cite{Abramsky2011}

A measurement scenario is defined as a triplet $\langle X, \mathcal{M}, O \rangle$, where $X$ refers to a collection of observables, $O$ is the possible outcomes, and $\mathcal{M}$ denotes an abstract simplicial complex composed of subsets from $X$. 

Every element in $X$ is an observable of the system under consideration. 
Upon measurement, each observable yields one of the outcomes contained in $O$.
The characterization of $\mathcal{M}$ as an abstract simplicial complex implies a particular structural feature: if a subset $C$ belongs to $\mathcal{M}$, then every subset nested within $C$ must also be an element of $\mathcal{M}$. 

A necessity of contextuality is that one cannot measure all the observables in $X$ simultaneously, at least not without altering the state of the system. 
Thus, every framework for contextuality must provide a description of the compatibility between observables.
Within the sheaf-theoretic framework, each simplex in the simplicial complex $\mathcal{M}$ constitutes a subset of observables in $X$ that are measurable simultaneously, i.e.\ they are mutually compatible. 
A \emph{measurement context}, or simply \emph{context}, is defined as a maximal simplex in $\mathcal{M}$, which is not a proper subset of any other simplex in $\mathcal{M}$.

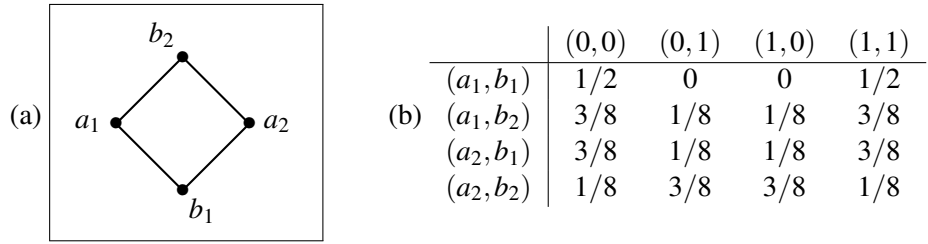
\begin{figure}
  \centering
  (a)
  \fbox{
  \begin{tikzpicture}[x=45pt,y=45pt,thick,label distance=-0.25em,baseline=(O.base), scale=0.7]
    \def\basenameone{$a_1$}
    \def\basenametwo{$b_1$}
    \def\basenamethree{$a_2$}
    \def\basenamefour{$b_2$}

    \def\outcomeone{$0$}
    \def\outcometwo{$1$}

    \coordinate (O) at (0,0);
    \def\sectionoffset{1.5}
    \def\sectionsep{0.5}

    \foreach \i in {0,1} {
        \coordinate (t\i) at (0, \sectionoffset + \sectionsep*\i);
    }

    \foreach \i/\p in {0/0,1/6,2/12,3/18} {
        \coordinate [inner sep=0em] (v\i) at ($ (
            {-cos(\p*pi/12 r)*0.8},
            {-sin(\p*pi/12 r)*0.8}
            ) $);
        \foreach \j in {0,1} {
            \coordinate [inner sep=0em] (v\i-\j) at ($ (v\i) + (t\j) $);
        }
    }

    \foreach \i/\j in {0/1,1/2,2/3,3/0} {\draw (v\i) -- (v\j);}


    \node [inner sep=0.1em,label={[label distance=-0.25em]left:{\basenameone}}] at (v0) {$\bullet$};
    \node [inner sep=0.1em,label={[label distance=-0.625em]330:{\basenametwo}}] at (v1) {$\bullet$};
    \node [inner sep=0.1em,label={[label distance=-0.25em]right:{\basenamethree}}] at (v2) {$\bullet$};
    \node [inner sep=0.1em,label={[label distance=-0.5em]175:{\basenamefour}}] at (v3) {$\bullet$};

\end{tikzpicture}
  }
  \qquad
  (b)
  \begin{tabular}{r|ccccc}
  & $(0, 0)$ & $(0, 1)$ & $(1, 0)$ & $(1, 1)$  \\ \hline
  $(a_1, b_1)$ & $1 / 2$ & $0$ & $0$ & $1 / 2$ \\
  $(a_1, b_2)$ & $3 / 8$ & $1 / 8$ & $1 / 8$ & $3 / 8$ \\
  $(a_2, b_1)$ & $3 / 8$ & $1 / 8$ & $1 / 8$ & $3 / 8$ \\
  $(a_2, b_2)$ & $1 / 8$ & $3 / 8$ & $3 / 8$ & $1 / 8$ \\
  \end{tabular}

  \caption{(a) The simplicial complex $\mathcal{M}$ in the Bell-CHSH scenario. 
  Every vertex represents an observable and every edge represents a context. 
  Alice chooses between $a_1$ and $a_2$; Bob chooses between $b_1$ and $b_2$. 
  The absence of edges between $a_1$ and $a_2$, and between $b_1$ and $b_2$, indicates their incompatibility. 
  (b) An empirical model of the Bell-CHSH scenario. Each row represents a joint probability distribution over the observables in the context. For example, the bottom-right entry $1/8$ is the probability of observing $a_2 = 1$ and $b_2 = 1$ when measuring the observables in the context $(a_2, b_2)$.}
  \label{fig:basechsh}
\end{figure}

For instance, the measurement scenario in the Bell-CHSH settings is specified by 
$X = \{a_1, a_2, b_1, b_2\}$; $\mathcal{M} = \big\{ \{a_1, b_1\}, \{a_1, b_2\}, \{a_2, b_1\}, \{a_2, b_2\} \big\}$; $O = \{0, 1\}$.
The simplicial complex $\mathcal{M}$ can be geometrically realized as the boundary of a square, where each vertex corresponds to an observable and each edge represents a context (see Figure~\ref{fig:basechsh}(a)).
Two parties are involved in this scenario: Alice is allowed to measure either $a_1$ or $a_2$, and Bob is allowed to measure either $b_1$ or $b_2$. The measurements are dichotomic, i.e.\ the outcomes are either $0$ or $1$.

Every subset of observables which is a context in $\mathcal{M}$ can be measured jointly. Thus we can define a (local) joint probability distribution over the observables in the context.
Such a joint probability distribution can either be estimated by performing the measurements in an experiment, or be calculated according to a theory of the system under consideration.
A collection of all such joint probability distributions is called an \emph{empirical model}, or simply \emph{model}, of the system.
For instance, using a set of appropriately chosen measurement bases, the Bell state $|\Psi \rangle = \big(|00\rangle + |11\rangle\big)/\sqrt{2}$ produces the empirical model depicted in Figure~\ref{fig:basechsh}(b). This state exhibits the highest violation of the Bell-CHSH inequality among all quantum states.

An empirical model is said to be \emph{signalling} if the marginalised distribution of a set of observables differs from one context to another.
In contrast, non-signalling implies that the observed probabilities remain invariant under different contexts, thereby preventing the transmission of information through the choice of context.

A prevalent misconception is a belief that \emph{signalling is contextuality}, often based on the incorrect reasoning that the \emph{probabilities} in a signalling model are generally context-dependent, leading to the conclusion that the model is contextual. 
However, it is essential to recognize a fundamental distinction between the two concepts: signalling pertains to the observed probabilities, while contextuality relates to the underlying hidden-variable theories of the model.

The qualitative criterion for contextuality of a model in the sheaf-theoretic framework is based on Fine's theorem~\cite{Fine1982}, which states that a model is contextual if and only if there exists a global probability distribution that is compatible with every local probability distribution in the model.

The quantitative degree of contextuality of a model is measured by the \emph{contextual fraction} {\cf}~\cite{Abramsky2017}.
Given an empirical model $e$, the contextual fraction $\text{\cf}(e)$ is defined as the minimum $\lambda$ such that $e$ admits a convex decomposition\footnote{Here, we represent the empirical models as empirical tables. Addition and scalar multiplication are then interpreted as standard matrix operations, where the empirical tables are treated as matrices.}:
\begin{equation}
  \label{eq:convexcf}
  e = (1-\lambda) e^{NC} + \lambda e^{C},
\end{equation}
where $e^{NC}$ is a non-contextual (and non-signalling) empirical model and $e^{C}$ is an empirical model that may be contextual. 

Suppose a given model $e$ is non-contextual, then $\lambda$ can be set to zero by choosing $e^{NC} = e$. Otherwise, $\lambda$ must be taken to be greater than zero to make the decomposition valid.
Therefore, for non-signalling models, the sheaf-theoretic criterion of contextuality is
\begin{equation}
  \label{eq:cfnosig}
  \text{\cf}(e) > 0.
\end{equation}
The calculation of {\cf} can be reduced to solving for a linear program, for which numerical solvers are readily available. The {\cf} of a model has a nice interpretation as the maximum amount of \emph{normalised violation} of all possible general Bell's inequalities~\cite{Abramsky2017}.

In the case of signalling models, the above decomposition cannot hold because $e^{NC}$ and $e^{C}$ are, by definition, non-signalling. We could consider allowing $e^{C}$ to be signalling. However, this adjustment would lead to the misleading conclusion that all signalling models are contextual, assuming we maintain our interpretation of {\cf} as a measure of contextuality for these models.

\subsection{Contextuality by Default}
In the setting of Contextuality-by-Default (CbD), there are two important notions: \emph{contents}, denoted by $q_i$, which are measurements, or more generally, questions about the system; and \emph{contexts}, denoted by $c^j$, which represent the conditions under which the questions are asked, e.g. their ordering. 
Every $q_i$ in a $c^j$ gives rise to a random variable $R^j_i$ taking values in $\{\pm 1\}$, and representing possible answers and their probabilities. All random variables in a given context are jointly distributed. 

A well-studied class of CbD systems are the cyclic systems~\cite{Dzhafarov2013,Dzhafarov2016a,Dzhafarov2015a}, where each context has exactly 2 contents and every content is in exactly 2 contexts.
The rank of a cyclic system is the number of contents, or equivalently, the number of contexts.

A cyclic system of rank $n$ is contextual if and only if $\cnt{1}$ is positive, where $\cnt{1}$ is defined as:
\begin{equation}\label{eq:BellInequality}
    \cnt{1} := s_{odd} \left(\left\{\left<R^{j}_{i_j}R^{j}_{i'_j}\right>\right\}_{j=1,\ldots,n}\right) - \Delta - n + 2 > 0
\end{equation}
where $i_j\neq i'_j$ for all $j$ and $R^{j}_{i_j}, R^{j}_{i'_j}$ are well-defined for all $j$. Quantities $s_{odd}: \mathbb{R}^n \to \mathbb{R}$ and $\Delta$ are defined as follows:
\begin{equation}
    s_{odd}\left(\underline{x}\right) = \max_{\substack{\underline{\sigma}\in \{\pm1\}^k; \\ \mathfrak{p}(\underline{\sigma}=-1)}}\underline{\sigma}\cdot \underline{x}\ ; \qquad 
\Delta = \sum_{i=1}^n \left|\left<R^{j_i}_{i}\right> - \left<R^{j'_i}_{i}\right>\right|
\end{equation}
where $\mathfrak{p}(\underline{\sigma}) = \prod_{i=1}^n \sigma_i$ ($\mathfrak{p}$ is the parity function of $\underline{\sigma}$).  
The quantity $\Delta$ measures the degree of signalling in the system. Thus, a non-signalling system has $\Delta=0$.

For a rank 4 cyclic system, i.e.\ the Bell-CHSH scenario, the above inequality reduces to the maximum violation of the Bell-CHSH inequalities over the choices of the four signs:
\begin{equation}
    \cnt{1} = \pm \left< R^0_0\ R^0_1 \right> \pm \left< R^1_1\ R^1_2 \right> \pm \left< R^2_2\ R^2_3 \right> \pm \left< R^3_3\ R^3_0 \right> - 2
\end{equation}
where the number of minus signs has to be taken odd. Therefore, the CbD criterion of contextuality coincides with the Bell-CHSH inequalities for the Bell-CHSH scenario.

\subsection{Ambiguous words as observables}
Ambiguities in natural language have posed a challenge to natural language processing.
Lexical ambiguity, where a word has multiple meanings, is one of the most common types of ambiguity in natural language. For instance, the word \textit{produce} has two possible meanings: \textit{to give birth} and \textit{to make something}. 

Without any context, it is not possible to determine which of the two meanings is intended.
Another type of ambiguity is \textit{coreference ambiguity}, where a word can potentially refer to different entities. For instance, the pronoun \textit{it} can refer to the \textit{dog} or the \textit{cat} in the sentence \textit{The dog chased the cat. It barked.}.
In this paper, we focus on the latter type of ambiguity.

A method to formalise the notion of contextuality in natural language is by viewing an ambiguous word as an observable, with its interpretations as possible outcomes.
For instance, the word \textit{produce} has (at least) two possible interpretations: \textit{to give birth} and \textit{to make something}.
Measuring the word \emph{produce} amounts to selecting one of these interpretations by a reader.

We can assign probabilities to these interpretations based on the frequency of the interpretations in an experiment where a group of readers is asked to interpret the word \textit{produce}, or a single reader is asked to assign a probability to each of the interpretations.
The first approach is more costly as it requires a large number of readers to be involved in the experiment.
However, the latter approach is better suited to machine learning models since they can be trained to assign probabilities to different interpretations.

This way of treating ambiguous words as observables was first proposed by Wang et al.~\cite{Wang2021a,Wang2021b}.
The authors considered subject-verb and verb-object phrases where each word carries at least two possible interpretations.
Measurement contexts were constructed by selecting different pairs of nouns and verbs, in a way similar to how Alice and Bob select their measurements in the Bell-CHSH scenario.
The probabilities in the results were estimated from a group of crowd workers who were asked to assign a score to the different interpretations.

\section{Winograd Schema Challenge}

Commonsense reasoning, the inherent human capacity to logically comprehend the world around us, has long been a focal point in the field of artificial intelligence, with the aim to cultivate this ability in machines.

The Winograd Schema Challenge (WSC) emerged as a measure of this commonsense reasoning capability. The challenge was inspired by Terry Winograd's seminal paper~\cite{Winograd1972}, wherein he contended that syntax alone falls short in the interpretation of natural language, necessitating commonsense or world knowledge as well. The challenge presents a collection of sentences, each with an ambiguous pronoun whose meaning can be clarified via the context. A machine is deemed to have passed the test if it can disambiguate the pronoun with an accuracy on par with human performance.

The classic example of a Winograd schema, originally constructed by Winograd himself, is the following pair of sentences:
\begin{exe}
  \ex \begin{xlist}
    \ex The city councilmen refused the demonstrators a permit because \textbf{they} \textit{feared} violence. 
    \ex The city councilmen refused the demonstrators a permit because \textbf{they} \textit{advocated} violence. 
  \end{xlist} 
\end{exe}
Note that the two sentences differ only in the words \textit{feared} and \textit{advocated}. In both sentences, there is an ambiguous pronoun \textbf{they} which can either refer to the \textit{city councilmen} or the \textit{demonstrators}. In the first sentence, it can be inferred through commonsense reasoning that the pronoun \textbf{they} refers to the \textit{city councilmen}, as it is within our common sense that city councilmen are the ones who tend to prevent violence in demonstrations. 
In the second sentence, the pronoun \textit{they} refers to the \textit{demonstrators}, as it is within our common sense (stereotype) that demonstrators tend to advocate violence and that doing so would lead to the refusal of a permit for a demonstration.

Another classic example of a Winograd schema is the following pair of sentences:

\textexample{
  The trophy doesn't fit into the suitcase because it's too [\textit{small} / \textit{large}].
}
Here we adopt a compact notation in which the pair of square brackets encloses the two possible word choices, each leading to a different sentence. This notation will be employed throughout the paper.

In a WSC, the participant is asked to identify the correct interpretation of the ambiguous pronoun.
Success in the test is defined by the participant's accuracy equalling or approximating human performance.
The evaluation of responses to a WSC question is straightforward, either the correct referent of the ambiguous pronoun is identified or not.

In contrast, the Turing Test has been criticised for being too difficult to evaluate.
Originated as the imitation game by Turing~\cite{Turing1950}, the test involves a human judge interrogating a machine via a textual interface. The conversation between the judge and the machine is unrestricted. If the judge or a panel of judges cannot distinguish the machine from a human based on the conversation, the machine is deemed to have passed the test.
However, this unrestricted nature of the Turing Test opens doors to potential deception. 
In fact, for a machine to pass the test, it must deceive as machines lack physical bodies. If questioned about its physical attributes, like height or weight, the machine must lie to successfully pose as a human.
Due to this advantage of the ease of evaluation over the Turing Test, the WSC was proposed as a replacement for the Turing Test.

Unlike the Turing Test, the WSC is a structured binary-choice test.
The major issue with the WSC is that it is over-constrained - it is unexpectedly difficult to construct examples of it, due to the numerous requirements that must be satisfied.
A valid Winograd schema must satisfy the following requirements:
\begin{enumerate}
  \item A Winograd Schema comprises a pair of sentences that differ slightly from each other. The first sentence includes a \textit{special} word which, when replaced by an \textit{alternate} word, yields the second sentence. For instance, in the \textit{trophy-suitcase} example, \textit{small} is the \textit{special} word, and \textit{large} is its \textit{alternate}.

  \item The sentences should contain two noun phrases. In the \textit{trophy-suitcase} example, \textit{the trophy} and \textit{the suitcase} serve as the two noun phrases.

  \item A pronoun, which agrees with the two noun phrases in number and gender, must be present in the sentences. For example, in the \textit{trophy-suitcase} scenario, the pronoun \textit{it} aligns with both \textit{the trophy} and \textit{the suitcase} regarding number and gender.

  \item The pronoun's referent should be easily identifiable from a natural reading of the sentence, and the correct referent should differ between the two sentences.

  \item Each sentence in the pair should be fluid and natural to read, to the extent that they could feasibly appear in regular text sources like news articles or Wikipedia pages.

\end{enumerate}
The outlined requirements ensure the preservation of both linguistic structure and the test's integrity:

\begin{enumerate}
\item The first requirement ensures grammatical consistency across the pair of sentences.
\item The fourth requirement necessitates a change in the correct referent of the pronoun when the special word is replaced with the alternate. This stipulation indicates that grammatical structure alone does not determine the correct pronoun referent.
\item The fifth requirement safeguards the authenticity of the language used in the test, ensuring it remains aligned with naturally occurring language.
\end{enumerate}
Crafting valid examples of the Winograd schema is a complex task due to the set restrictions and requirements. The challenge of creating such schemas is evidenced by the limited number of examples in the original Winograd Schema Challenge set, which includes only 285 instances\footnote{Available at \href{https://cs.nyu.edu/davise/papers/WinogradSchemas/WS.html}{https://cs.nyu.edu/davise/papers/WinogradSchemas/WS.html}.}. 

In 2018, the first system achieved a better-than-chance accuracy of 57.1\%~\cite{Emami2018} on the original 285 examples of the WSC. In 2019, a fine-tuned RoBERTa~\cite{Liu2019} model achieved a human-like accuracy of 90.1\%~\cite{Sakaguchi2021}.
The WSC has suffered from the same problem that plagued the Turing Test -- there are weaknesses in the test that can be exploited without having to demonstrate the desired human-level intelligence. Simply put, the WSC has been defeated~\cite{Kocijan2022}. 

It is even more so for the WSC precisely because of its ease of evaluation. Proposals to increase the difficulty of the WSC, such as requiring the test-taker to select a correct explanation for their answer from a list of options~\cite{Zhang2020,He2021}, emerged as potential solutions. However, these suggestions further complicate the already challenging task of question set construction. An alternative could involve requiring free-form explanations from the test-taker, though this would likely introduce additional ambiguity and make the evaluation process more difficult.

\section{Generalised Winograd Schema}
In this section, we present our approach for the generalisation of the Winograd Schema, enabling the potential observation of contextuality. We will first discuss why the original Winograd Schema is insufficiently complex to exhibit contextuality, and then propose a generalised Winograd Schema that is sophisticated enough to host contextuality.

\subsection{Modelling Winograd Schemas as measurement scenarios}
To study the contextuality in the Winograd Schema, we model it with a measurement scenario in the sheaf-theoretic framework. This way of treating ambiguity in language is akin to the way ambiguous phrases are treated in~\cite{Wang2021a}, where an ambiguous word is considered an observable in a measurement scenario.

However, the same ambiguous word, i.e.\ the ambiguous pronoun, is shared across the twin pair of sentences in a Winograd Schema.
Thus, if we follow the approach of ``words as observables'' strictly, then we will end up with a trivial measurement scenario, where there is only one observable, i.e.\ the ambiguous pronoun. 
Moreover, this naive approach deviates from the spirit of the Winograd Schema, which is to disambiguate a pronoun by considering the linguistic context. 
Instead, We argue that there should be exactly two contexts in the measurement scenario, one for each sentence in the twin pair. 
Recall that in the original Winograd Schema, the twin pair of sentences are identical except for the special word and the alternate word. In a rough sense, the special word and the alternate word provide the \textit{linguistical context} for disambiguating the pronoun. 
This way of defining the measurement contexts provides a concrete link between \textit{context in language} and \textit{contextuality in quantum mechanics}.

Following from the above discussion, we define an observable as a tuple: (\textbf{pronoun}, \textit{special word}) or (\textbf{pronoun}, \textit{alternate word}), to distinguish between the two pronouns in different linguistical contexts.
The possible outcomes of each of the two observables are the candidate referents of the pronoun.

\begin{definition}[Winograd Schema scenario]
  Given a Winograd Schema with two noun phrases A and B; an ambiguous pronoun \textbf{p} which refers to either A or B; a special word (\textit{s}) and an alternate word (\textit{a}),
  the corresponding measurement scenario is defined by the data:
  \begin{itemize}
    \item observables $X = \{ (\textbf{p}, \textit{s}), (\textbf{p}, \textit{a}) \}$;
    \item contexts $\mathcal{M} = \bigl \{ \{(\textbf{p}, \textit{s})\}, \{(\textbf{p}, \textit{a})\} \bigr \}$;
    \item outcomes $O = \{\text{A}, \text{B}\}$.
  \end{itemize}
  We call such a measurement scenario a \emph{Winograd Schema scenario}, or a WS scenario in short.
\end{definition}

With the \textit{councilmen-demonstrators} example, the measurement scenario would be given by the data:
\begin{itemize}
  \item observables $X = \{ (\textbf{they},\ \textit{feared}),\ (\textbf{they},\ \textit{advocated}) \}$;
  \item contexts $\mathcal{M} = \bigl \{ \{(\textbf{they},\ \textit{feared})\},\ \{(\textbf{they},\ \textit{advocated})\} \bigr \}$;
  \item outcomes $O = \{\text{city councilmen},\ \text{demonstrators}\}$.
\end{itemize}
It becomes apparent that any Winograd Schema scenario is too simplistic to accommodate any contextual model due to the absence of overlapping contexts.
One can always construct a compatible global distribution by taking the product of the local distributions.

\subsection{Generalising the Winograd Schema scenario}
Before proceeding to the generalisation of Winograd Schema, we point out an interpretation of the WS scenario as an analogy to an experiment in quantum physics.
Consider an imaginary experimenter, Alice, who decides whether to measure the pronoun with the special word, or with the alternate word. That is, Alice chooses between the two observables: $(\textbf{p}, \textit{s})$ and $(\textbf{p}, \textit{a})$.
This is exactly analogous to Alice choosing between two projection axes in an experiment measuring a spin-1/2 particle.

A natural and obvious way to generalise the WS scenario would be to add one more experimenter, Bob. This results in the Bell-CHSH scenario, which is well-known to be able to host contextual models.
That amounts to introducing one more pronoun, one more special word and its alternate word, to the original Winograd Schema.
We use the subscript $1$ to denote objects relating to the first pronoun and the subscript $2$ to denote objects relating to the second pronoun.

Here we give a set of requirements for the generalised Winograd Schema, in the style of the original WSC: 
\begin{enumerate}
  \item A generalised schema consists of four slightly differing sentences. The first sentence contains two special words $\textit{s}_1$ and $\textit{s}_2$. Similar to the original Winograd Schema, $\textit{s}_1$ can be replaced by an alternate word $\textit{a}_1$ and $\textit{s}_2$ can be replaced by an $\textit{a}_2$.
  The possibility of replacing special words with alternate words creates the rest of the four sentences.
  \item There are a pair of noun phrases.
  \item There are two pronouns in the sentences. The first pronoun refers to one of the noun phrases in the first pair of noun phrases. The second pronoun refers to either one noun phrase in the second pair of noun phrases.
  \item All four sentences should be natural to read.
\end{enumerate}
In short, a generalised Winograd Schema is two Winograd Schemas put together in a single discourse.

\begin{definition}[Generalised Winograd Schema scenario]
  Given a Generalised Winograd Schema with two noun phrases A and B; two ambiguous pronouns \textbf{p}$_1$ and \textbf{p}$_2$ can each refers to either A or B; two special words (\textit{s}$_1$) and (\textit{s}$_2$); two alternate words (\textit{a}$_1$) and (\textit{a}$_2$), the corresponding measurement scenario is defined by the data:
\begin{itemize}
  \item observables $X = \{(\textbf{p}_1, \textit{s}_{1}), (\textbf{p}_1, \textit{a}_{1}), (\textbf{p}_2, \textit{s}_{2}), (\textbf{p}_2, \textit{a}_{2})\}$
  \item contexts $\mathcal{M} = \bigl\{ \{ (\textbf{p}_1, \textit{s}_{1}), (\textbf{p}_2, \textit{s}_{2}) \}, \{ (\textbf{p}_1, \textit{s}_{1}), (\textbf{p}_2, \textit{a}_{2}) \}, \{ (\textbf{p}_1, \textit{a}_{1}), (\textbf{p}_2, \textit{s}_{2}) \}, \{ (\textbf{p}_1, \textit{a}_{1}), (\textbf{p}_2, \textit{a}_{2}) \}\bigr\};$
  \item outcomes $O = \{\text{A}, \text{B}\}$.
\end{itemize}
Such a measurement scenario is called a \emph{Generalised Winograd Schema scenario}, or a generalised WS scenario in short.
\end{definition}
The generalised WS scenario is isomorphic, i.e.\ identical upon relabelling, to the Bell-CHSH scenario shown in Figure~\ref{fig:basechsh}. It has long been known that the Bell-CHSH scenario can host contextual models~\cite{Bell1964,Clauser1969}. Thus a carefully designed generalised Winograd Schema would be able to demonstrate contextuality. 

Here we provide a straightforward example of a generalized Winograd Schema scenario, built upon the original \textit{trophy-suitcase} example:
\textexample{ The trophy doesn't fit into the suitcase because \textbf{it}$_1$ is too [\textit{s}$_1$ = \textit{small} / \textit{a}$_1$ = \textit{large}]. Nonetheless, \textbf{it}$_2$ is [\textit{s}$_1$ = \textit{light} / \textit{a}$_2$ = \textit{heavy}].}
The corresponding generalised WS scenario is given by:
\begin{itemize}
  \item observables $X = \{(\textbf{it}_1, \textit{small}), (\textbf{it}_1, \textit{large}), (\textbf{it}_2, \textit{light}), (\textbf{it}_2, \textit{heavy})\}$
  \item contexts $\mathcal{M} = \begin{aligned} \bigl\{ &\{ (\textbf{it}_1, \textit{small}), (\textbf{it}_2, \textit{light}) \}, \{ (\textbf{it}_1, \textit{small}), (\textbf{it}_2, \textit{heavy}) \},\\ &\{ (\textbf{it}_1, \textit{large}), (\textbf{it}_2, \textit{light}) \}, \{ (\textbf{it}_1, \textit{large}), (\textbf{it}_2, \textit{heavy}) \}\bigr\};\end{aligned}$
  \item outcomes $O = \{\text{trophy}, \text{suitcase}\}$.
\end{itemize}
Interestingly, it was in the original set of Winograd Schemas (WSC285) that Davis designed a special example making use of two pronouns: 
\textexample{Sid explained his theory to Mark but \textbf{he} couldn't [\textit{convince} / \textit{understand}] \textbf{him}.}
The author deemed this example a ``Winograd schema in the broad sense'' since using more than one pronoun violates the requirements of the original Winograd Schema.
Yet, this example is not a proper generalised Winograd Schema defined in this paper, as it only employs one special word and one alternate word.

Other than the fact that its scenario is too simple, there is another reason why the original Winograd Schema is not contextual: the intended referent of the pronoun should be obvious to a human reader.
That means an empirical model constructed with judgement data collected from human subjects on the original Winograd Schema would be deterministic or nearly deterministic. 
It is known that deterministic systems are not contextual~\cite{Dzhafarov2019}.
On the other extreme, a completely random model is trivially non-contextual.
Intriguingly, it seems that only a system with a moderate level of intelligence, in between that of humans and that of complete randomness, would have the possibility of being contextual. 

There are two directions to where we could take the generalised Winograd Schema: (1) to continue its mission to be a test of intelligence or commonsense reasoning; (2) to become a well-structured linguistic setting under which contextual models could be found. 

Recent results from large language models have demonstrated human-like accuracies in solving the Winograd Schema Challenge.
The introduction of one more pronoun might increase the difficulty of the challenge, possibly stipulating advancements in the field of natural language processing.
However, it is our goal to find bridges between natural language and contextuality. Therefore the second direction will be the focus of this paper.

\subsection{An example of the generalised Winograd Schema}
As our goal is to uncover contextual models in natural language, we need to gather judgment data from human participants to build empirical models for generalized Winograd Schema instances. 
Crucially, deterministic systems lack contextuality. Therefore, our generalized Winograd Schema examples should be inherently ambiguous to human readers, unlike the original Winograd Schema where humans can easily resolve the pronoun.

Due to the requirement of having two almost identical pairs of naturally-sounding sentences, it is a difficult task to come up with examples of the original Winograd Schema. 
The extra requirements we put forward for the generalised Winograd Schema make it even harder to come up with naturally-sounding examples.
Here we report an example of the generalised Winograd Schema\footnote{It was pointed out by one of the reviewers that the original version of the example contains several incorrect uses of English. Here we provide the corrected version of the example.}:
\textexample{
  A and B belong to the same [\textit{cannibalistic} / \textit{herbivorous}]$_1$ species of animal. On a hot afternoon in the south Sahara, \textbf{one of them}$_1$ was very hungry. They noticed each other when they were roaming in the field. After a while, \textbf{one of them}$_2$ is no longer [\textit{hungry} / \textit{alive}]$_2$.
}
Note that we had to violate the requirement of having a single sentence because it is difficult to come up with a naturally-sounding sentence that contains every ingredient of the generalised Winograd Schema.
We also decided to use the referring phrase \textbf{one of them} instead of the third-person pronoun \textbf{it} to improve the naturalness of the example.

We used the alphabetic symbols A and B as the two noun phrases as we wanted to make the two symmetric. 
That is, any empirical model of the scenario is invariant to the interchanging of A and B.
It turns out that all symmetric models are non-signalling, at least for cyclic scenarios such as that Bell-CHSH scenario. 
Dealing with symmetric models carries two disadvantages: (1) it is more difficult to assert the contextuality of a signalling model; (2) the sheaf-theoretic criterion of contextuality applies to non-signalling models only.
By considering only symmetric models, we thereby avoid the complications of dealing with non-signalling models.

\subsection{Human judgements on the example}
\begin{figure}[h]
  \centering
  \fbox{
  \includegraphics[width=0.95\textwidth, trim=0.5cm 21cm 0.5cm 0.5cm, clip]{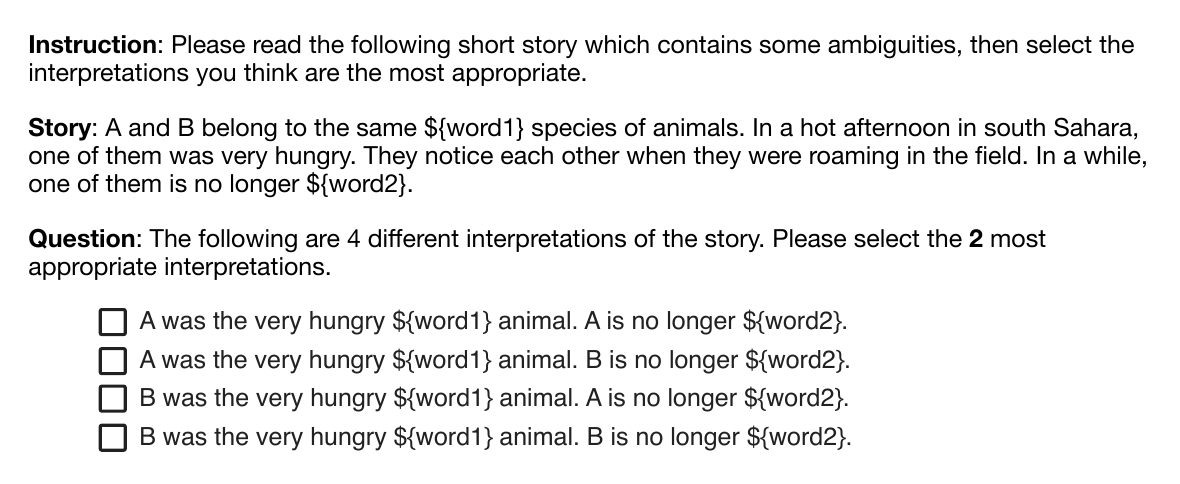}
  }
  \caption{A screenshot of the template of the questionnaire. The placement holders \textsf{\$\{word1\}} and \textsf{\$\{word2\}} are instantiated with the two special words or the alternate words of the generalised Winograd Schema. In this example, \textsf{\$\{word1\}} can be either \textit{cannibalistic} or \textit{herbivorous} and \textsf{\$\{word2\}} can be either \textit{hungry} or \textit{alive}. Four versions of the questionnaire were created, each corresponding to one of the four contexts in the generalised WS scenario. 
  \emph{Note that the story contains several incorrect uses of English. Unfortunately, we did not notice these until a reviewer pointed them out, after data collection.}}
  \label{fig:questionnaire}
\end{figure}

We collected human judgments on this example on the crowd-sourcing platform Amazon Mechanical Turk in the form of a questionnaire. 
There were four versions of the questionnaire, each corresponding to one of the four contexts in the generalised WS scenario.
The respondents were asked to read the example and answer a question about the correct referents, A or B, of the two referring phrases \textbf{one of them}$_1$ and \textbf{one of them}$_2$.
A screenshot of the questionnaire is shown in Figure~\ref{fig:questionnaire}.

Since each referring phrase can be interpreted in two ways, there are 4 possible combinations of interpretations, (A, A), (A, B), (B, A), (B, B), of the two referring phrases.
The symmetry between A and B in the example ensures that the combinations (A, A) and (B, B) are equally plausible and (A, B) and (B, A) are also equally plausible.
Therefore we asked the respondents to pick two out of the four combinations.
This design choice also allows the detection of invalid answers, that is, those that do not respect the symmetry between A and B.

A total of 410 responses were collected on Amazon Mechanical Turk separately on two dates: 20th Oct 2022 and 23rd Nov 2022.
Out of the 410 responses, 110 were to the context (\textit{cannibalistic}, \textit{hungry}) and 100 each were to the rest of the three contexts.
Out of all the responses, 348 were valid, i.e.\ their responses respected the symmetry between A and B. The respondents were each financially rewarded USD~1.00, regardless of the validity of their responses.

The collected valid data were used to build an estimated probability distribution for each of the four contexts. The resulting empirical model is shown in Table~\ref{table:sahara}.
The model violates the Bell-CHSH inequality by 0.192 with a standard deviation of 0.176.
Since the model is symmetric in the outcomes by construction, it is non-signalling and thus the measure of contextuality \cnt{1} in the CbD framework coincides with the degree of violation~\cite{Kujala2019}.
The symmetry in the outcomes also allows the violation to saturate the bound defined by {\cf} in sheaf-theoretic framework~\cite{Abramsky2017}, i.e.\ the following equality is attained
\begin{align}
\max\left\{0, \frac{1}{2}\ \textsf{violation of Bell-CHSH inequality} \right\} = \cf.
\end{align}
Thus, our model is considered contextual in both the sheaf-theoretic framework and the CbD framework.

\begin{figure}[h]
  \centering
  \includegraphics[width=0.6\textwidth]{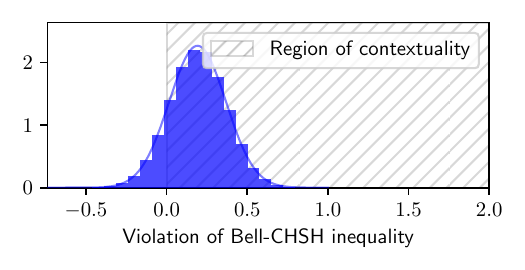}
  \caption{
  A normalised histogram of the Bell-CHSH inequality violation for 100,000 bootstrap samples from the model shown in Table~\ref{table:sahara}. A positive violation, indicative of contextuality, is observed in 87\% of the resampled models.  
  The standard deviation of the distribution is 0.176. }
  \label{figure:histo}
\end{figure}

To establish the significance of the contextuality result, we conducted bootstrap resampling to estimate the spread of the violation to the Bell-CHSH inequality.
Simulated datasets were generated by random sampling with replacement from the original dataset. The resulting distribution of violations is depicted in Figure~\ref{figure:histo}. 
Among the resampled datasets, 87\% of them exhibited a positive violation, indicating that our experimental model demonstrates contextuality with a significance level of 87\%.

\begin{table}
\centering  

\begin{tabular}{ll|ccccc}
  (a) && (A, A) & (A, B) & (B, A) & (B, B)  \\ \hline
  (\textit{canni},& \textit{hungry}) & 0.402 & 0.097 & 0.097 & 0.402 \\
  (\textit{canni},& \textit{alive}) & 0.044 & 0.455 & 0.455 & 0.044 \\
  (\textit{herbi},& \textit{hungry}) & 0.345 & 0.154 & 0.154 & 0.345 \\
  (\textit{herbi},& \textit{alive}) & 0.344 & 0.155 & 0.155 & 0.344 \\
\end{tabular}
\begin{tabular}{l|ccccc}
  (b) & (A, A) & (A, B) & (B, A) & (B, B)  \\ \hline
\dots & $1/2$ & $0$ & $0$ & $1/2$ \\
\dots & $0$ & $1/2$ & $1/2$ & $0$ \\
\dots & $1/2$ & $0$ & $0$ & $1/2$ \\
\dots & $1/2$ & $0$ & $0$ & $1/2$ \\
\end{tabular}
\caption{(a) The empirical model constructed with the 410 human judgments collected from Amazon Mechanical Turk. 
The violation of Bell's inequality of the model is 0.192~$\pm$~0.176. 
For brevity, the special word \textit{cannibalistic} is shortened to \textit{canni} and the alternate word \textit{herbivorous} is shortened to \textit{herbi}.
The model generally resembled the PR model shown in Table (b) on the right.
}
\label{table:sahara}
\end{table}

\section{Conclusions and Future Work}
In this work, we employed the sheaf-theoretic framework for contextuality to model the Winograd Schema, originally formulated as an ambiguous coreference resolution task. Our findings revealed that the original Winograd Schema scenario lacked the necessary complexity to exhibit contextuality. To address this limitation, we introduced an additional ambiguous pronoun and a new pair of special and alternate words, creating a generalized Winograd Schema reminiscent of the Bell-CHSH scenario. Through crowdsourcing, we collected human judgments on an example of the generalized Winograd Schema and observed a contextual empirical model with a significance level of 87

An intriguing direction for future research involves constructing a comprehensive set of examples based on the proposed generalized Winograd Schema, thereby establishing it as a new challenge in the field of natural language processing.
One potential approach is to leverage state-of-the-art generative language models such as GPT-4 to systematically generate examples of the schema with minimal human intervention.
Careful prompt engineering would be needed to ensure that the generated examples are of high quality.

As collecting human judgments is costly and time-consuming, another alternative approach for constructing empirical models of the generalized Winograd Schema involves utilizing generative language models to generate responses to examples. 
This approach also offers an opportunity to explore the extent to which the responses generated by language models align with human responses. By comparing and analysing the correspondence between model-generated responses and human responses, one could gain insights into the capabilities and limitations of language models in capturing the way human beings understand language.

This paper presents an approach that consists of deliberately constructing sentences that exhibit contextuality. 
This strategy of ``detecting contextuality in natural language'' may invite criticism for its contrived nature.

An alternative approach could involve the application of mathematical frameworks designed for contextuality to analyze pre-existing natural language data, moving away from the intentional construction of examples with distinct features \cite{Wang2023}. 
The aim of this strategy would not be to pursue contextuality within natural language. Instead, it would focus on developing novel methods for modelling natural language phenomena from a different perspective.

\section*{Acknowledgements}
We are grateful to Daphne Wang for insightful discussions and the anonymous reviewers for their constructive comments.
KL is supported by the Engineering and Physical Sciences Research Council [grant number EP/S021582/1].
MS is supported by the Royal Academy of Engineering research chair RCSRF2122-14-152 on Engineered Mathematics for Modelling Typed Structures.
\bibliographystyle{eptcs}
\bibliography{references}

\end{document}